# Recognizing Multiple Ingredients in Food Images Using a Single-Ingredient Classification Model

Kun Fu[a], and Ying Dai[a,] *

[a]*Iwate Prefectural University, Sugo 152-52, Takizawa and 020-0693, Japan*

A B S T R A C T

Recognizing food images presents unique challenges due to the variable spatial layout and shape changes of ingredients with different cooking and cutting methods. This study introduces an advanced approach for recognizing ingredients segmented from food images. The method localizes the candidate regions of the ingredients using the locating and sliding window techniques. Then, these regions are assigned into ingredient classes using a convolutional neural network (CNN)-based single-ingredient classification model trained on a dataset of single-ingredient images. To address the challenge of processing speed in multi-ingredient recognition, a novel model pruning method is proposed to enhances the efficiency of the classification model. Subsequently, the multi-ingredient identification is achieved through a decision-making scheme, incorporating two novel algorithms.

The single-ingredient image dataset, designed in accordance with the "新食品成分表 food 2021" (New Food Ingredients List FOODS 2021), encompasses 9,982 images across 110 diverse categories, emphasizing variety in ingredient shapes. In addition, a multi-ingredient image dataset is developed to rigorously evaluate the performance of our approach. Experimental results validate the effectiveness of our method, particularly highlighting its improved capability in recognizing multiple ingredients. This marks a significant advancement in the field of food image analysis.

## 1. Introduction

In recent years, China's obesity epidemic has intensified significantly. According to a report, over half of the adult population in China is overweight, a situation attributed to unhealthy lifestyles, poor eating habits, and the consumption of high-energy and high-fat foods [1]. Food, which is fundamental to human survival, consistently influences our lives. The capacity to recognize various food types and their ingredients is crucial for numerous health-related analyses, such as calorie estimation and nutritional breakdown. This knowledge is vital for dietary management, particularly for individuals who are health-conscious.

In recent years, the field of food ingredient recognition has increasing attention due to its potential applications in various industries. This includes its utilization in buffet restaurants, where such technology can automatically monitor food consumption and estimate caloric content.

However, identifying ingredients and dishes presents several challenges. The first challenge is diversity: the ingredient can appear various shapes based on the way it is cut and cooked. The second challenge is the similarity among some ingredients, as different ingredients can look very similar. The third challenge is overlapping and mixing in food images, where ingredients may overlap or mix each other. The fourth issue is the lack of standardized definition for food ingredients in public datasets. Lastly, the sheer variety of dishes is innumerable.

Current research in food ingredient identification primarily falls into two categories. The first involves the use of Multi-label learning for ingredient identification, as seen in works [2], [3], and [4]. Works [2] and [3] apply Multi-Task Learning and Region-Wise Recognition to the domain of food ingredient identification, while work [4] constructs a Multi-Attention Network (IG-CMAN) for this purpose. However, the use of Multi-label learning has limitations: building a dataset using Multi-labels is time-consuming and labor-intensive, and there are issues of ingredients interference with each other in recognition. The second category employs semantic segmentation for ingredient identification, as demonstrated in works [5], [6], [7] and [19]. Work [5] introduces a framework named [FoodSAM], which combines coarse semantic masks with masks generated by SAM. Work [6] proposes a multi-modal pre-learning method named ReLeM for semantic segmentation, and work [7] introduces a method called Feature-Suppressed Contrast (FeaSC) to enhance self-supervised food pre-training. Work [19] presents a compact and fast multi-task network, namely FoodMask, for clustering-based food instance counting, segmentation and recognition. Like Multi-label learning, creating a Pixel-Wise annotation dataset for Semantic segmentation is also a time-consuming and laborious process. Furthermore, if ingredients are not correctly segmented, the method relying solely on Pixel Regions for identification is not adequate. Hence, work [8] presents a framework using a single ingredient classification model to segment ingredients in food images to solve the challenge of overlapping and mixing regarding ingredients. However, this work focuses only on ingredient segmentation and not on identification. It also faces challenges such as incomplete separation of ingredients and different ingredients being segmented as the same.

In this paper, we propose a novel system for multiple ingredient recognition from food images. This study introduces an advanced approach for recognizing ingredients segmented from the food images. The method localizes the candidate regions of the ingredients using the locating and sliding window techniques. Then, these regions are assigned into ingredient classes using a CNN (Convolutional Neural Network)-based single-ingredient classification model trained on a dataset of single-ingredient images. To address the challenge of processing speed in multi-ingredient recognition, a novel model pruning method is proposed to enhance the efficiency of the classification model. Subsequently, the multi-ingredient identification is achieved through a decision-making scheme, incorporating two novel algorithms. The contributions of our paper can be summarized as follows:

1) A framework for recognizing multiple ingredients in food images is proposed, featuring two decision-making algorithms to enhance performance.

2) A hierarchical single-ingredient image dataset is constructed based on the food standard taxonomy to train single-ingredient classification model. The dataset includes as many images with different cutting methods and cooking styles as possible, addressing the challenge of diversity in the same ingredient.

3) A new method for pruning models is proposed, which effectively reduces model parameters and enhances model speed. Additionally, it's found that the pruned model also works well for ingredient segmentation.

---

* Corresponding author. Tel.: +081-19-694-2544; fax: +081-19-694-2544; e-mail: dai@iwate-pu.ac.jp

## 2. Related work

In this study, we utilize the framework proposed in [8] to segment the ingredients in the food images. However, some challenges arise with the above segments, such as separating some ingredients into different segments and grouping different ingredients together as the same segment.

Moreover, we employ the techniques of Making Decision [13,14], Transfer Learning [11], and pruning [12]. Making Decision refers to the process of making wise choices in specific situations. In the technical field, it is commonly used to determine the final output results. Transfer Learning is a machine learning approach that utilizes already trained models and knowledge and transfers them to new tasks and domains. Pruning is employed to reduce the model parameters, thereby making the network lighter and improving the inference speed. A commonly used pruning method is Taylor pruning, which defines the pruning criterion based on the Taylor series expansion of the loss function and prunes the filters of convolutional layers based on this criterion.

On the other hand, regarding public datasets on food ingredients, there are FoodSeg103 [6], VIREO FOOD-251 [15], and ISIA FOOD200 [16]. Since the VIREO FOOD-251 dataset currently lacks food ingredient labels, and the ISIA FOOD200 dataset is not available, testing on these two datasets is not feasible. Therefore, FoodSeg103 is used as the public dataset for evaluating the performance of the proposed method afterwards.

## 3. Dataset

To construct the new dataset for ingredient recognition, we define 110 ingredient categories based on the '新食品成分表 food 2021' (New Food Ingredients List FOODS 2021) in [9] and [10]. These ingredients are listed in Figure 1, covering a wide range of categories defined in the standard food taxonomy.

### 3.1 Single ingredient image dataset (SI110)

We collect image samples of individual ingredients, aiming to capture a diverse range of shapes. For instance, our dataset includes images of eggs in various forms such as boiled, scrambled, and fried, shown in Figure 2. We ensure that 5-10 samples with similar appearance are collected for each ingredient to mitigate bias towards a certain ingredient in a certain appearance.

Some ingredients exhibit numerous appearance variations due to cutting and cooking conditions, while others have fewer variations. Consequently, this diversity results in a long-tail distribution within the dataset for ingredient categories, as depicted in Figure 3. This distribution is instrumental in preventing class-bias during model training.

On the other hand, to address the potential issue of misidentifying the background as an ingredient, we introduce a background class with approximately 100 background images in the dataset.

SI110 dataset contains 9982 ingredient images, involving in 110 kinds of ingredients. This dataset is used to train the single-ingredient classification model.

| 1 | abalone | 31 | crab | 61 | octopus | 91 | soybean |
|---|---|---|---|---|---|---|---|
| 2 | almond | 32 | cream | 62 | okra | 92 | spinach |
| 3 | apple | 33 | cucumber | 63 | onion | 93 | squids |
| 4 | asparagus | 34 | daikon | 64 | orange | 94 | strawberry |
| 5 | avocado | 35 | eel | 65 | other white flesh | 95 | sunflower seed |
| 6 | bamboo shoot | 36 | egg | 66 | oyster mushroom | 96 | sweat potato |
| 7 | banana | 37 | eggplant | 67 | oysters | 97 | seaweed |
| 8 | bean sprout | 38 | enoki | 68 | papaya | 98 | swine |
| 9 | bitter melon | 39 | fig | 69 | peach | 99 | tofu |
| 10 | black rice | 40 | garlic stem | 70 | peanuts | 100 | tomato |
| 11 | blueberry | 41 | grape | 71 | pear | 101 | tree ears |
| 12 | bok choy | 42 | grape fruits | 72 | peas | 102 | tuna |
| 13 | bonito | 43 | green soybean | 73 | pecan | 103 | wakame |
| 14 | broad beans | 44 | green onion | 74 | pepper | 104 | watermelon |
| 15 | broccoli | 45 | hazel nuts | 75 | pineapple | 105 | wax gourd |
| 16 | cabbage | 46 | kidney bean | 76 | pistachio | 106 | wulnuts |
| 17 | carrot | 47 | kidney beans | 77 | pitaya | 107 | wheat_product |
| 18 | cashews | 48 | konpu | 78 | potato | 108 | yam |
| 19 | cattle | 49 | kiwi | 79 | poultry | 109 | yogurt |
| 20 | cauliflower | 50 | lemon | 80 | pumpkin | 110 | yellow peach |
| 21 | celery stem | 51 | lettuce | 81 | pumpkin seeds | | |
| 22 | celtuce | 52 | lobster | 82 | purple laver | | |
| 23 | cheese | 53 | lotos | 83 | raspberries | | |
| 24 | cherry | 54 | mackerels | 84 | rice | | |
| 25 | chestnuts | 55 | mango | 85 | salmon | | |
| 26 | chickpea | 56 | mantis shrimp | 86 | sesame seeds | | |
| 27 | chinese cabbage | 57 | melon | 87 | shiitake | | |
| 28 | chinese chieves | 58 | millet | 88 | shimeiji | | |
| 29 | clam | 59 | mushroom | 89 | shrimp | | |
| 30 | corn | 60 | meat_product | 90 | snowpea | | |

Fig. 1. 110 kinds of ingredients

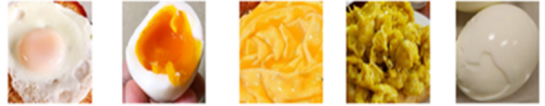

Fig. 2. Examples of ingredient samples

### 3.2 Multi-ingredient food image dataset (MIFI)

This dataset is used to evaluate the performance of multiple ingredient recognition in food images. MIFI dataset contains a total of 2121 images. Each image contains multiple ingredients. Two examples are shown in Figure 4. The distribution of how many ingredients is included in each image is shown in Figure 5.

The MIFI dataset includes all the ingredients defined in the SI110 dataset. From the distribution in Figure 5, we can see that currently the MIFI contains the most images with 2 or 3 kinds of ingredients.

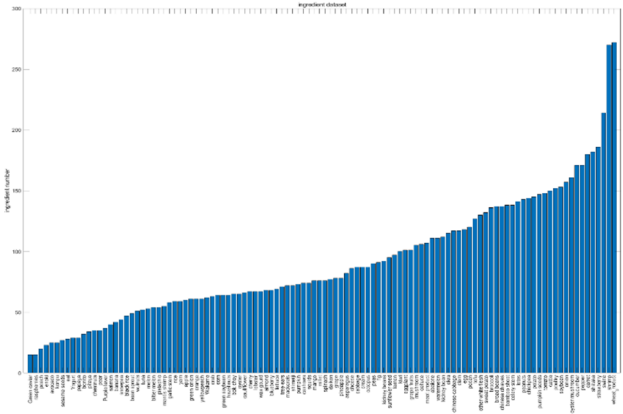

Fig. 3. The distribution of ingredient quantity

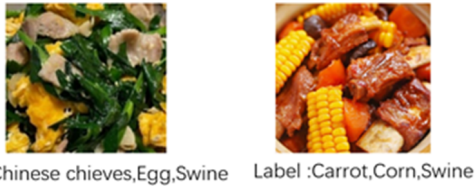

Fig. 4. Examples from MIFI dataset

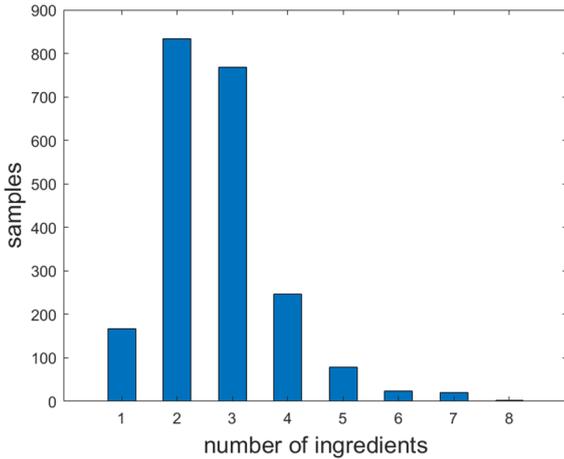

Fig. 5. The distribution of how many ingredients is included in each image

## 4 Methodology

In this section, we propose a framework for the multiple ingredient recognition system, as shown in Figure 6. The system localizes candidate regions from segments for recognition using the locating and sliding window techniques. These regions are then assigned to ingredient classes using a convolutional neural network (CNN)-based single ingredient recognition model trained on the SI110 dataset. Finally, the ingredients in the segments are identified using a decision-making scheme based on these candidate results.

The segmentation is based on the framework proposed in [8]. Initially, the feature maps of the image are extracted through the last CNN layer of the single ingredient classification model. Subsequently, K masks are generated via Pixel-wise K-means clustering based on these feature maps. Here, 'K' represents the number of clusters. There are currently two methods to obtain the value of 'K'. One is through the information of the image annotation, and the other is automatic prediction [18]. Finally, the ingredients are segmented based on the generated masks.

Then, the segmented images are fed into the single ingredient classification model for recognition. As demonstrated in Figure 6, if the segment is inputted into the model without any processing, the recognition outcome is inaccurate, primarily because the background occupies most of the image area weakening the information about the ingredient. Consequently, to better identify the ingredients, we employ the locating and sliding window techniques to extract the ingredients candidate regions in the segments.

The method of localizing candidate regions of the ingredients is explained in the following. An example is shown in Figure 7.
1) Erosion operation [17] is used to eliminate small and insignificant objects from the segmented images.
2) Dilation operation is used to fill the cavities in the object and eliminate small particle noise in the target object.
3) Localized boxes are generated by measuring properties of the image regions.

Sliding window is a common method in object detection where the system scans over the image in small patches (windows), checking for the presence of ingredients, as shown in Figure 8. In this paper, the window length and width are 1/3 of the image length and width respectively, while the horizontal and vertical strides are also 1/3 of the image dimensions.

In the following, we introduce how to build a single-ingredient classification model and use it to recognize the ingredients from the candidate regions in detail.

*4.1 Single-ingredient classification model and its pruning*

We use SI110 as the training dataset and conduct transfer learning based on EfficientNetB0 [12] and EffcientNetB3 to train a single-ingredient classification model.

Given the repetitive invocation of the single-ingredient classification model through the sliding window method, we integrate model pruning techniques to augment the system's recognition speed. Among the prevalent pruning methods, Taylor pruning is commonly utilized, which defines an index for pruning derived from the Taylor expansion of the loss function.

In this study, we propose a novel pruning method for the CNN model. The feature maps within CNN blocks are reflective of the structural parameters embedded in the convolutional layers, with the similarity between feature maps indicative of analogous structural parameters. We posit that a substantial similarity in the sums of feature maps between any two blocks suggests a diminished role for one of the blocks. Consequently, we present a new method for pruning the network's blocks, involving the calculation of a similarity matrix between each pair of blocks, followed by the removal of one of the two most similar blocks. The pruning procedure is illustrated in Figure 9. The detailed explanation is provided below.
1. Initiate the activation of the CNN model to acquire feature maps.
2. Evaluate the similarity between every pair of blocks by computing the Structural Similarity Index Measure (SSIM) based on the cumulative sum of feature maps from each block.
3. Identify the two most similar blocks and subsequently remove one of them based on the SSIM values.

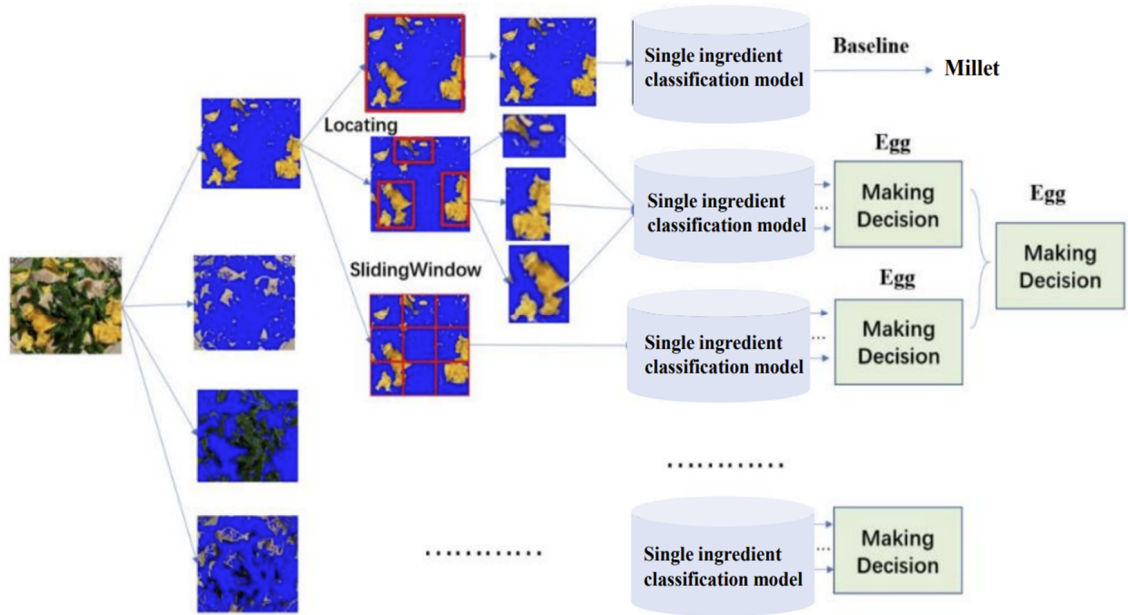

Fig. 6. Multi-ingredient identification system

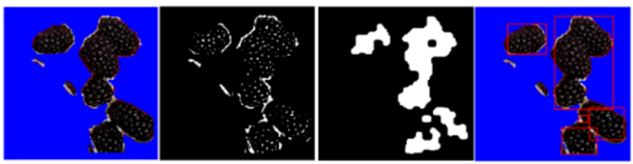

Fig. 7. Locating the candidate regions

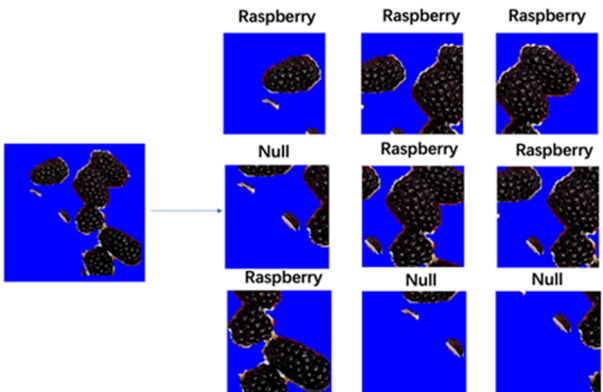

Fig. 8. Sliding Window diagram

4. Implement fine-tuning on the pruned CNN to uphold its performance.
5. Iteratively traverse the process of comparing blocks and remove one of the two most similar blocks.
6. Determine whether to proceed with pruning based on the SSIM values. If the maximal value in the SSIM matrix falls below a predetermined threshold, cease the pruning process.

This pruning approach focuses on enhancing the efficiency of the model by reducing the number of redundant blocks, thereby streamlining the network. By opting to remove an entire block rather than just individual filters, this approach achieves a more notable reduction in both parameters and computational complexity. This can be particularly beneficial in scenarios where more efficient model is required, such as in environments with limited computational resources.

Here is an example of the pruning process applied to the EfficientNetB0 model. For a given image, the sums of feature maps from blocks 1 to 15 of EfficientNetB0 are illustrated in Figure 10.

The SSIM matrices between each pair of blocks in the first round of the pruning process are presented in Figures 11. From Figure 11, it can be observed that block 12 and block 13 exhibit the highest similarity in feature maps. Consequently, in accordance with our proposed pruning method, block 13 is removed. This process is iteratively repeated for subsequent pruning rounds, and after each pruning iteration, fine-tuning is conducted, followed by the recalculation of the SSIM matrix based on the updated model.

Fig. 9. Pruning flow chart

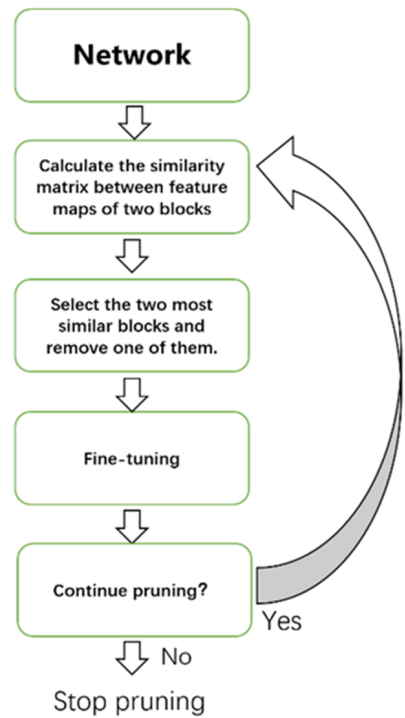

As the structure of the convolutional layer within the block used for extracting feature maps remains invariant, the feature maps derived by this layer for any given image should demonstrate similarity. Consequently, utilizing a single image is enough for calculating the similarity matrix between each pair of blocks concerning their feature maps.

However, it is imperative to highlight that the elemental values within the similarity matrix undergo changes when alterations are made to the input image. The magnitude of these variations is intricately tied to the complexity inherent in the image. Therefore, to determine the optimal threshold for the pruning process, it becomes indispensable for each task to compute the similarity matrix using an alternative image that effectively encapsulates the distinctive characteristics germane to that task.

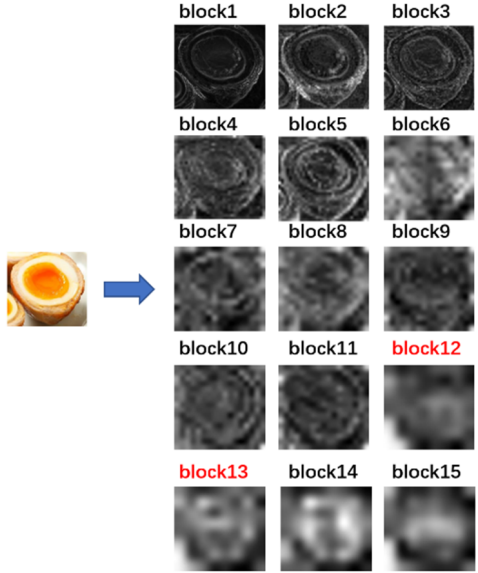

Fig. 10. Feature map sums from block 1 to 15 of EfficientNetB0

Fig. 11. First pruning feature map and similarity matrix between each pair of blocks

*4.2 Making decision*

Upon preliminary examination, it is found that neither the locating nor the sliding window, when employed independently, yields satisfactory results for ingredient recognition. The locating method alone is deemed unsuitable for certain images due to limitations in the segmentation process, as detailed in [8], which may result in multiple ingredients being segmented into the same image. Conversely, relying solely on the sliding window method for recognition is impeded by the varying sizes of ingredients, presenting challenges in setting an appropriate window size. As a remedy, we design two decision-making algorithms for ingredient identification that integrate the results obtained from candidate regions extracted through both the locating and sliding window methods.

Algorithm 1 assumes that each segment produces only one output. The flowchart is depicted in Figure 12. Initially, the image is input into the system. Subsequently, the two methods of locating ingredient regions and sliding window operate concurrently to identify areas in the image that may contain ingredients. These regions are fed into the single-ingredient classification model, resulting in two lists of potential ingredients along with their class scores. For each list, in cases where the number of every ingredient are different, a majority vote determines the correct ingredient.

Conversely, the ingredient with the highest average score is selected. Ultimately, the results from both lists are amalgamated to make the final decision regarding the ingredient, based on the highest average scores. Figure 13 illustrates a specific instance of Algorithm 1 in action.

To address the issue of multiple ingredients being segmented into the same segment, we further develop Algorithm 2 for their recognition. In Algorithm 2, more than one results can be outputted. The flowchart of this algorithm is shown in Figure 14. Same as Algorithm 1, the image is input into the system. Then, two methods of locating ingredient regions and sliding window operate simultaneously to identify areas in the image that may contain ingredients. These regions are fed to the single-ingredient classification model, resulting in two lists of potential ingredients along with their corresponding class scores. Additionally, we introduce a top-n approach, wherein the ingredients with the highest scores, ranking in the top n, are selected from the two lists. Subsequently, the same selected ingredients are merged. Ultimately, m (where m ≤ n) ingredients are determined as the recognition results.

This approach not only mitigates the issue of multiple ingredients being segmented into the same segment but also ensures that only the most confidently recognized ingredients are considered in the final decision.

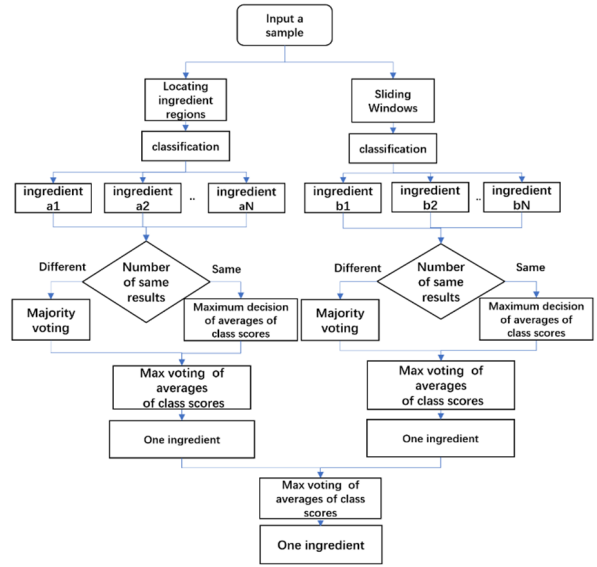

Fig. 12. Algorithm 1 flow chart

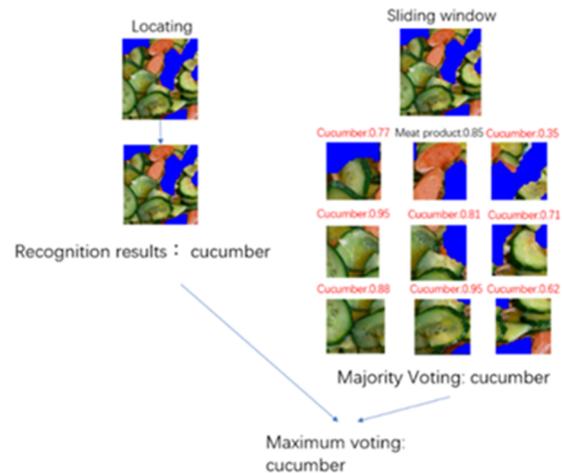

Fig. 13. An example of Algorithm 1

Figure 15 illustrates an example of Algorithm 2 with a top 5 selection. As depicted in the figure, each identified region is assigned a class score. From these, the top 5 candidates, indicated by red characters, are selected based on their scores and subsequently merged if their names are identical. As a result, cucumber and meat product are recognized.

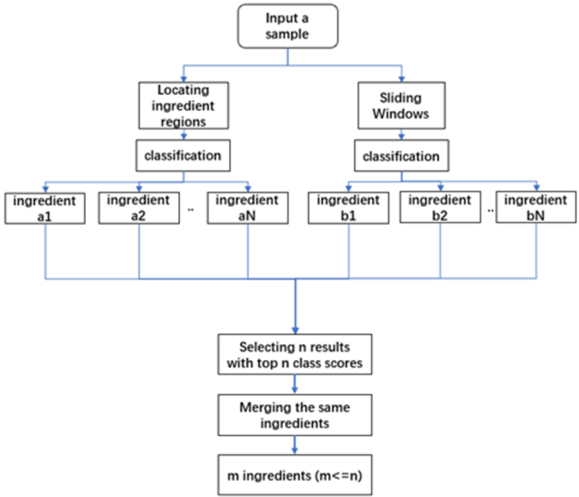

Fig. 14. Algorithm 2 flow chart

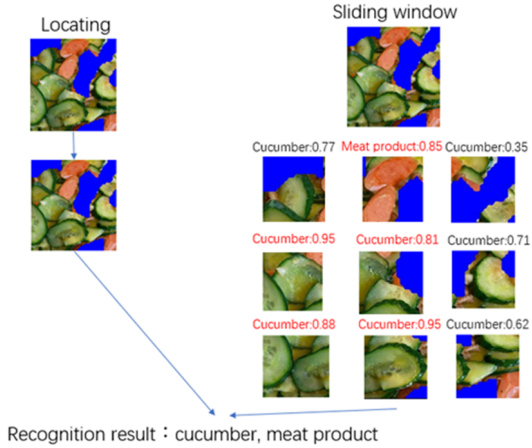

Fig. 15. An example of Algorithm 2

## 5. Experiments and analysis

The experiments are implemented on a computer with the specifications of GPU: NVIDIA GeForce RTX 4080 Laptop GPU 12GB, Memory 64GB, and are implemented on MATLAB 2023b.

### 5.1 Evaluation metrics

We use the following metrics for evaluating the performance of the proposed methods. The accuracy denotes a rate of correctly assigned samples to all samples, precision measures the correctness of positive predictions, while recall assesses the system's ability to capture all relevant instances. The F1 Score is the harmonic mean of precision and recall.

$$\text{accuracy} = \frac{\Sigma_{i=1}^{N} TP_i}{Total\ Samples} \quad (1)$$

$$precision_i = \frac{TP_i}{TP_i + FP_i} \quad (2)$$

$$recall_i = \frac{TP_i}{TP_i + FN_i} \quad (3)$$

$$F1Score_i = 2 \times \frac{Precision_i \times Recall_i}{Precision_i + Recall_i} \quad (4)$$

Where, *i* denotes a certain class or a certain image, *N* does the number of classes or images.

### 5.2 Experimental results and analysis

A. Segmentation

Firstly, let us introduce ingredient segmentation. Our experimental findings indicate that the pruned single-ingredient classification model produces highly satisfactory segmentation results. As depicted in Figure 16, the top row illustrates segmentation outcomes using the original EfficientNetB0, which are evidently unsatisfactory. The second row showcases segmentation results obtained with the fine-tuned EfficientNetB0 on the SI110 dataset. The third row exhibits segmentation results derived from the pruned fine-tuned EfficientNetB0. It is discernible that the pruned model demonstrates superior segmentation performance.

We postulate that this superiority arises from the fact that the EfficientNetB0, pretrained on ImageNet, tends to overly emphasize texture features while neglecting color and shape characteristics, resulting in a texture feature bias. Furthermore, as the images in the SI110 dataset prominently feature textures, the fine-tuned model retains an inclination towards texture. However, pruning the fine-tuned model effectively mitigates this tendency.

Accordingly, in the subsequent sections, we utilize the segments generated by the EfficientB0-based pruned model to assess its performance in recognizing multiple ingredients in food images.

On a parallel note, employing a pruned model for feature extraction in segmentation offers an additional advantage: it enhances processing speed when compared to the original model (EfficientNetB0). This efficiency is particularly conspicuous in our experiments, where we conducted segmentations on images with a resolution of 1100×1360. Through 10 segmentation trials on the image, followed by the computation of the average speed, we observe that utilizing the original model for segmentation consumes an average of 2.73 seconds. In contrast, the pruned model requires only 1.95 seconds, on average, for segmenting the image.

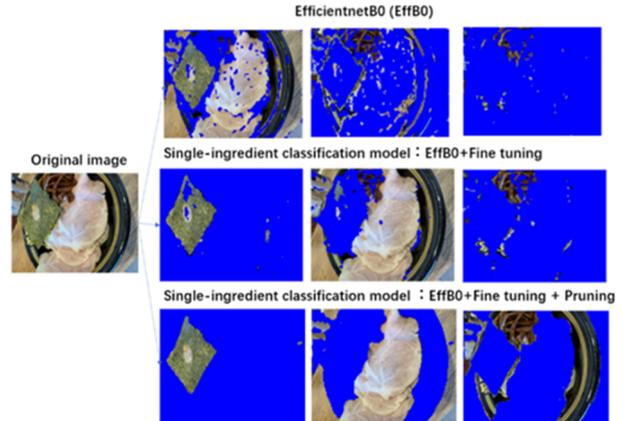

Fig 16. Examples of segmentation using different models

B. Evaluating single-ingredient classification model

Table 1. Single-ingredient classification Model performance on SI110 Dataset

| Model | Batch size | Accuracy | Avg precision (%) | Avg recall (%) | Avg f1Score (%) |
|---|---|---|---|---|---|
| EfficientNetB0-based | 5 | 67.72 | 58.19 | 59.45 | 57.25 |
|  | 8 | 68.46 | 66.47 | 67.15 | 64.98 |
|  | 16 | 70.08 | 69.72 | 69.39 | 67.88 |
|  | 32 | 74.36 | 78.57 | 81.48 | 76.92 |
|  | 64 | 68.23 | 69.60 | 69.01 | 67.65 |

Table 1 displays the experimental results of the single-ingredient classification model fine-tuned on the SI110 dataset using EfficientNetB0. Examining the table reveals that varied batch sizes lead to diverse experimental outcomes. Presently, the optimal result is attained with a batch size of 32, achieving an accuracy of 74.36% and an average F1 score of 76.92%.

C. Error analysis

We conducted an error analysis based on the confusion matrix regarding the classification results of individual ingredients. All ingredients with an F1 Score below 0.5 were extracted, as depicted in Figure 17. Upon scrutinizing the results, it is evident that the low recognition performance stems from the high inter-similarity of certain ingredients. For instance, the low score for beef is attributed to its strong visual resemblance to pork after cooking, leading to confusion between the two. Similarly, eggs, melon and Squids face challenges due to their visual similarities to pumpkin, cheese, and octopus in certain culinary preparations. The close phylogenetic relation between mushroom and oyster mushroom, both being fungi, also contributes to their visual resemblance.

Conversely, ingredients such as yogurt, cream, and yam exhibit poor performance owing to the lack of discriminative features.

Fig 17. Error analysis based on the Confusion Matrix

D. Evaluating pruned single-ingredient classification model

We additionally evaluated the performance of the pruned model, and the test results are presented in Table 2. The pruning process was based on the EfficientNetB0-based fine-tuned model (EffB0+FT), which was trained with the batch size of 32 on the SI110 dataset. Examining the table, in comparison to the original model, the third-round pruned model, despite encountering a 4% decline in accuracy, demonstrated a reduction in size by approximately one-third and an increase in speed by one-fourth.

Table 2. Pruned model performance on SI110 dataset

| Model | Proposed | Speed(S) | Parameter(KB) | Accuracy(%) |
|---|---|---|---|---|
| EffB0+FT | Original model | 0.04 | 15422 | 74.36 |
| | First pruning | 0.04 | 13260 | 72.55 |
| | Second pruning | 0.04 | 11097 | 71.34 |
| | Third pruning | 0.03 | 10318 | 70.43 |
| | Fourth pruning | 0.03 | 9919 | 68.52 |

*Note: EffB0+FT (fine tuning)): EfficientNetB0-based fining tuning (batch size=32) model

We also conducted a comparison between our proposed pruning method and the Taylor pruning method [11]. As illustrated in Table 3, our third-round pruning model, in contrast to Taylor pruning, sustains a same level of accuracy while achieving a reduction in parameter size by more than double. Moreover, the speed has experienced an increase of more than threefold.

Table 3. Comparing our proposed pruning method with the Taylor pruning

| Model | Method | Speed | Parameter(KB) | Accuracy(%) |
|---|---|---|---|---|
| EffB3+FT | TP | 0.10 | 26333 | 70.02 |
| EffB0+FT | BP | 0.03 | 10318 | 70.43 |

*Note: Taylor pruning: TP (Taylor pruning)
Proposed pruning: BP (Ours, block pruning)

E. Evaluating Algorithm 1 and Algorithm 2

We assess the performance of Algorithm 1 and Algorithm 2 for multiple ingredient recognition in food images using the MIFI dataset. The EffB0+FT model, incorporating the proposed block pruning (EffB0+FT+BP), is employed for segmenting ingredient candidates from the images, and the EffB0+FT is utilized for classifying the ingredients within the candidate regions.

Table 4-a and Table 4-b present the precisions, recalls, and F1Scores achieved by Algorithm 2 under various top-n settings. The results in Table 4-a are calculated based on class-wise metrics, while those in Table 4-b are derived from image-wise metrics. From the findings in the tables, it is observed that both average and median precisions exhibit a decrease with an increase in the value of n, while the trend in recall is precisely the opposite. The average F1Score, as an overall measurement of precision and recall, attains two maximal values corresponding to the top-1 and top-2 settings. Hence, if both F1Score and recall are emphasized, the top-2 setting for Algorithm 2 (Algorithm 2 (top2)) is deemed most suitable.

Table 4-a. Performance of algorithm 2 with different top n (class-wise)

| Algorithm2 | Median & Average | Precision (%) | Recall (%) | F1Score (%) |
|---|---|---|---|---|
| Top1 | Average | 51.79 | 44.99 | 43.03 |
| | Median | 52.23 | 43.26 | 42.03 |
| Top2 | Average | 43.71 | 51.63 | 42.19 |
| | Median | 42.76 | 53.84 | 40.57 |
| Top3 | Average | 38.30 | 56.42 | 40.80 |
| | Median | 36.13 | 58.06 | 41.50 |
| Top5 | Average | 30.90 | 61.33 | 36.56 |
| | Median | 28.14 | 63.63 | 36.84 |

Table 4-b. Performance of algorithm 2 with different top n (image-wise)

| Algorithm2 | Median & Average | Precision (%) | Recall (%) | F1Score (%) |
|---|---|---|---|---|
| Top1 | Average | 45.21 | 55.33 | 47.13 |
| | Median | 50.00 | 50.00 | 50.00 |
| Top2 | Average | 41.65 | 58.38 | 46.27 |
| | Median | 40.00 | 50.00 | 50.00 |
| Top3 | Average | 37.73 | 61.10 | 44.46 |
| | Median | 33.33 | 50.00 | 44.44 |
| Top5 | Average | 30.14 | 66.76 | 39.68 |
| | Median | 30.14 | 66.67 | 40.00 |

To further demonstrate the effectiveness of Algorithm 1 and Algorithm 2, we assess their performance in a way of ablation study. These include a baseline approach (directly feeding the segmented image into EffB0+FT) and methods commonly utilized in existing research for localizing target regions. These methods encompass sliding window solely on the original image, sliding window solely on the segmented image, and locating candidate regions on the segmented image. Subsequently, Algorithm 2 (top2) is applied to the localized regions.

The plot graphs of class-wise precision, recall, and F1Score of Algorithm 1, Algorithm 2, and other methods are shown in Figure 18, Figure 19 and Figure 20. For Algorithm 2, the results are obtained under the top-2 setting.

In Figures 18, 19, and 20, the vertical axes represent the values of average precision, average recall, and average F1-score, respectively, utilizing the EffB0+FT model. The horizontal axis denotes the methods employed for recognition.

Examining Figure 18, it is evident that both the average and median precisions of the method, which exclusively localizes candidate regions on the segmented image with Algorithm 2 (top2), surpass those of Algorithm 1 and Algorithm 2. However, as depicted in Figure 19, the average and median recalls of Algorithm 2 are the highest, with the average recall being approximately 14% higher than the one of the aforementioned methods.

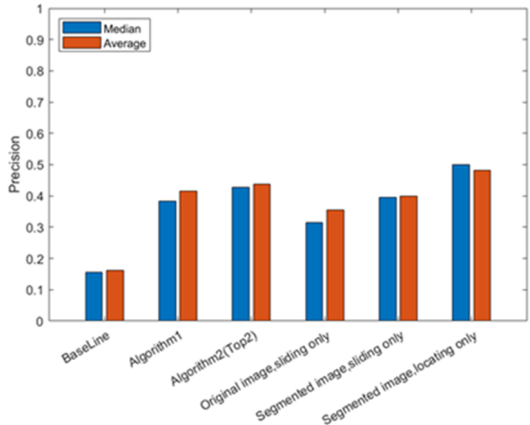

Fig 18. Precision results of various methods

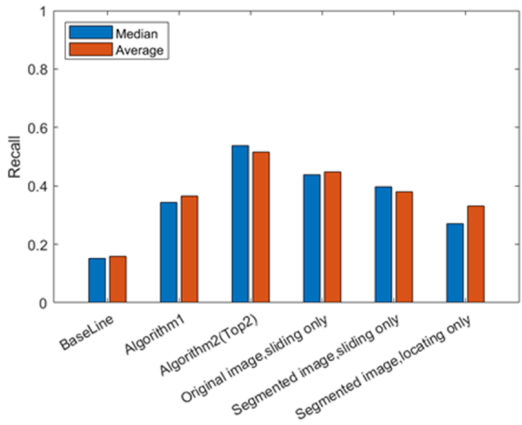

Fig 19. Recall results of various methods

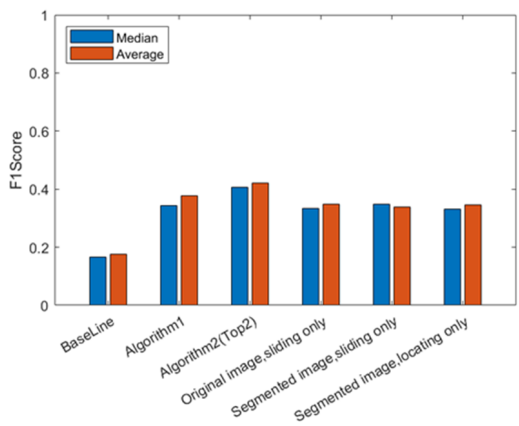

Fig 20. F1Score results of various methods

Furthermore, the average recall of Algorithm 2 (top2) is approximately 15.13% higher than that of Algorithm 1. This discrepancy arises from the fact that Algorithm 1 retains only one recognition result, whereas Algorithm 2 has the capability to output multiple recognition results. While this approach may result in some incorrect outcomes, it contributes to a lower precision yet a higher recall for Algorithm 2.

Figure 20 presents the results for the F1 Score, a crucial metric that amalgamates precision and recall to offer a balanced measure of a model's accuracy. Examining the outcomes in Figure 20, it is apparent that both Algorithm 1 and Algorithm 2 significantly outperform the baseline method, as well as other methods solely employing sliding window or region localization on segmented images. The average F1 Score of Algorithm 2 surpasses the baseline by 25% and exceeds the other methods by approximately 9%. Furthermore, Algorithm 2 demonstrates superior performance compared to Algorithm 1. This performance difference can be attributed to existing segmentation challenges, where different ingredients are not entirely separated. As Algorithm 1 outputs only one result, there is a possibility of missing ingredients.

Additionally, it is noteworthy that the disparity between the mean and median values of precision, recall, and F1 Score is minimal across all methods. This observation indicates that the results are not skewed towards particular ingredient categories. Consequently, it can be deduced that the outcomes of the EffB0+FT model, trained on SI110, demonstrate negligible class bias.

F. Comparing with SOTA methods

Additionally, we compare the proposed methods with state-of-the-art (SOTA) methods on the public dataset FoodSeg103, and the results are presented in Table 5. Since FoodSeg103 and our SI110 share 35 identical ingredient categories, we can calculate the average recall of these 35 categories, which is equivalent to the mAcc of other methods in Table 5.

In our results, "Ours_1" signifies our method where segments are obtained by the EffB0+FT+BP model, and the ingredients are recognized by Algorithm 2 using the EffB0+FT model. Similarly, "Ours_2" represents our method with segments obtained from the EffB0+FT+BP model, and ingredient recognition is performed by Algorithm 2 using the EffB0+FT+BP model. The "top2" and "top5" notations denote the settings of Algorithm 2. It is noteworthy that the models of references [5-7] were trained and verified on the same dataset as FoodSeg103 that is the dataset with the pixel-level annotation, showcasing results from the independent and identically distributed (i.i.d) setting, while our method was trained on our own dataset SI110 that is the dataset with the instance-level annotation, and verified on FoodSeg103, demonstrating results in the out-of-distribution (o.o.d) setting.

From the results in Table 5, it is observed that for our proposed method, the mAcc (average recall) using the pruned model is approximately 3% lower than that using the unpruned model.

On the other hand, although the results of Ours_1(top2), Ours_2(top2), and Ours_2(top5) are slightly lower than the methods of [5] [6], Ours_1(top5) achieves the highest mAcc (average recall) of 60.66% among all the methods. Considering our results are derived from an out-of-distribution (o.o.d) scenario, and the marginal difference between the results in Table 4 on the MIFI dataset and those in Table 5 on the Foodseg103 dataset, our proposed method demonstrates greater competitiveness and stronger generalization ability. Moreover, our ingredient classification model has been trained on an instance-level dataset, which is less time-consuming for annotation. This approach enhances the efficiency of the training procedure.

Furthermore, the approach outlined in [19] achieves an F1 (image-wise) of 60.81%. However, due to the overlap of 35 identical ingredient categories between FoodSeg103 and our SI110, our method cannot attain this performance on FoodSeg103. Nonetheless, the maximum F1 score (image-wise) achieved by our method is 47.13% on MIFI dataset, which is 13.68% lower than [19]. Beyond the out-of-distribution issue, our method exhibits weakness in accurately recognizing ingredients with high inter-class similarity. Enhancing the performance in identifying similar ingredients is imperative for future endeavors.

Table 5. Comparison with state-of-the-art (SOTA) performance

| Method | mAcc (average recall) (%) | F1(image-wise) |
|---|---|---|
| FoodSAM [5] | 58.27 | |
| SimSiam+FeaSC [7] | 46.20 | |
| CCNet-Finetune [6] | 53.80 | |
| ReLeM-CCNet-Finetune [6] | 59.50 | |
| FoodMask (ImageNet) [19] | | 60.81 |
| Ours_1(top2) | 53.06 | |
| Ours_1(top5) | 60.66 | |
| Ours_2(top2) | 50.06 | |
| Ours_2(top5) | 57.84 | |

*Note:
Ours_1: seg. By EffB0+FT+BP, reg. by EffB0+FT+Algorithm2
Ours_2: seg. by EffB0+FT+BP, reg. by EffB0+FT+BP+Algorithm2

G. Visualization of recognizing process

Figure 21 illustrates the process of ingredients recognition. The input image displays a dish containing mixed ingredients, including carrots, tree ears, and yams. On the right-hand side, the processed results are presented, showcasing the ability of Algorithm 2 to identify individual ingredients even when they are not fully separated.

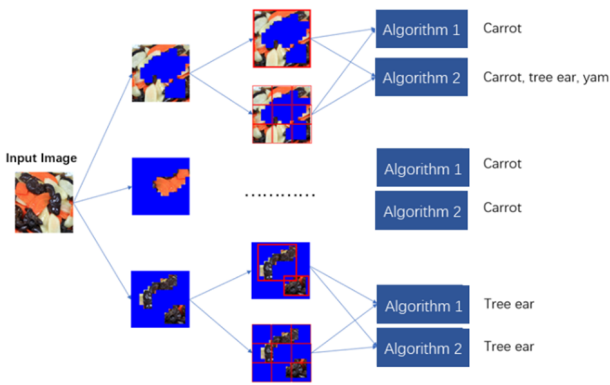

Fig 21. An example of ingredients recognition process

## 6. Conclusion

In this paper, we introduced a novel framework and developed a system for the recognition of multiple ingredients in food images. The proposed framework utilized a single-ingredient classification model for segmenting ingredients in food images and subsequently identifying them. Additionally, we built a comprehensive standard dataset, encompassing a wide range of standardized ingredients, excluding categories such as fats, desserts, and beverages. To address the challenge of variability within the same ingredient, we extensively collected data reflecting different cutting and cooking methods for each ingredient.

Furthermore, a new pruning method was proposed, demonstrating not only an acceleration of the single-ingredient classification model but also promising outcomes in ingredient and food segmentation. Two algorithms were designed, employing a decision-making scheme that integrated the classification results obtained from candidate regions using locating and sliding window techniques. Experimental results validated the effectiveness of the proposed method. In comparison with state-of-the-art methods, our approach demonstrates competitiveness and robust generalization capabilities.

**Acknowledgments**

This work was partly supported by JSPS KAKENHI Grant Number JP22K12095.